\documentclass{llncs}
\usepackage{makeidx}  
\usepackage{graphicx, siunitx, dblfloatfix, caption, subcaption, multirow, color, cite}
\begin{document}
\frontmatter          
\pagestyle{headings}  

\mainmatter              
\title{Segmentation of Intracranial Arterial Calcification with Deeply Supervised Residual Dropout Networks}
\titlerunning{Segmentation of Intracranial Calcification}  
%
\author{Gerda Bortsova\textsuperscript{1}, Gijs van Tulder\textsuperscript{1}, Florian Dubost\textsuperscript{1}, Tingying Peng\textsuperscript{2}, Nassir Navab\textsuperscript{2,3}, Aad  van der Lugt\textsuperscript{4}, Daniel  Bos\textsuperscript{4,5,6}, Marleen  de Bruijne\textsuperscript{1,7}
}
\authorrunning{} 
%
\tocauthor{}

\institute{
\textsuperscript{1} Biomedical Imaging Group Rotterdam, Erasmus MC, The Netherlands \\
\textsuperscript{2} Computer Aided Medical Procedures, Technische Universit{\"a}t M{\"u}nchen, Germany \\
\textsuperscript{3} Computer Aided Medical Procedures, Johns Hopkins University, USA \\
\textsuperscript{4} Department of Radiology and Nuclear Medicine, Erasmus MC, The Netherlands \\
\textsuperscript{5} Department of Epidemiology, Erasmus MC, The Netherlands \\
\textsuperscript{6} Department of Clinical Epidemiology, Harvard T.H. Chan School of Public Health, USA \\
\textsuperscript{7} Department of Computer Science, University of Copenhagen, Denmark
}

\maketitle              

\begin{abstract}
Intracranial carotid artery calcification (ICAC) is a major risk factor for stroke, and might contribute to dementia and cognitive decline.
Reliance on time-consuming manual annotation of ICAC hampers much demanded further research into the relationship between ICAC and neurological diseases.
Automation of ICAC segmentation is therefore highly desirable, but difficult due to the proximity of the lesions to bony structures with a similar attenuation coefficient.
In this paper, we propose a method for automatic segmentation of ICAC; the first to our knowledge. 
Our method is based on a 3D fully convolutional neural network that we extend with two regularization techniques.
Firstly, we use deep supervision to encourage discriminative features in the hidden layers.
Secondly, we augment the network with skip connections, as in the recently developed ResNet, and dropout layers, inserted in a way that skip connections circumvent them.
We investigate the effect of skip connections and dropout.
In addition, we propose a simple problem-specific modification of the network objective function that restricts the focus to the most important image regions and simplifies the optimization.
We train and validate our model using 882 CT scans and test on 1,000.
Our regularization techniques and objective improve the average Dice score by 7.1\%, yielding an average Dice of 76.2\% and 97.7\% correlation between predicted ICAC volumes and manual annotations.
\end{abstract}

\section{Introduction}

Intracranial arteriosclerosis has been established as a major cause of stroke \cite{bos2014intracranial} and might contribute to the risk of cognitive impairment and dementia \cite{bos2012atherosclerotic}.
Intracranial carotid artery calcification (ICAC) is a reliable marker for intracranial arteriosclerosis \cite{bos2014intracranial}.
ICAC lesions are identified in non-contrast computed tomography (CT) images as groups of voxels with an attenuation coefficient above 130 Hounsfield units (HU) on the track of the intracranial internal carotid artery (IICA), from its petrous part until the circle of Willis.

Further investigation into causes and consequences of ICAC might result in development of new treatments and preventive measures.
For example, ICAC volume, i.e., the total volume of all ICAC lesions found in a patient, might potentially be used in stroke risk estimation in clinical practice.

Automated ICAC segmentation is challenging for several reasons.
Firstly, identifying the IICA location requires information from a large neighborhood, due to a lack of contrast between arteries and surrounding tissues.
Secondly, ICAC might be very close to bones, which have similar intensity. (Refer to Fig. \ref{fig:det} for examples.)

To our knowledge, no methods have been proposed for automatic detection of ICAC.
However, a number of methods exist to automatically detect calcifications in other vessel beds \cite{van2014atherosclerotic, shahzad2012automatic}.
Perhaps the most well-studied problem is coronary artery calcification (CAC) detection.
Earlier automatic CAC scoring methods use supervised classification, but rely on atlas-based coronary artery localization \cite{wolterink2016evaluation}.
More recently, a deep learning approach \cite{wolterink2016automatic} was proposed to detect CAC in a more end-to-end fashion.
However, the close proximity of ICAC to bones makes its detection a different problem than detection of calcifications in cardiac or extracranial carotid arteries, as lesions there are usually relatively easy to distinguish from their immediate surroundings (the artery lumen).

Recently, deep neural networks demonstrated state-of-the-art performance on many challenging visual recognition tasks \cite{lecun2015deep}. Fully convolutional networks (FCNs) achieved impressive results for segmentation of both natural \cite{long2015fully} and biomedical images \cite{ronneberger2015u}.
FCNs are by design more computationally efficient and have higher capacity for accurate localization than patch-based approaches (e.g., \cite{ciresan2012deep}). Compared to purely convolutional nets without downsampling layers (e.g., the aforementioned CAC detection network \cite{wolterink2016automatic}), FCNs allow for much more features and/or layers and are hence capable of capturing more complex patterns.

Overfitting is a notorious problem of deep networks.
One of the ways to counter it is to use dropout layers \cite{srivastava2014dropout}, which introduce noise in hidden layers during training.
Another problem of deep networks is challenging optimization.
To combat it, Lee et al. \cite{lee2015deeply} proposed supervision of hidden layers, or ``deep supervision''.
This technique was reported to improve the convergence speed and reduce overfitting by encouraging the network to develop features useful for final classification in earlier layers.
To ease the optimization of very deep networks, He et al. \cite{he2016deep} introduced the residual network (ResNet) architecture.
ResNet is composed of blocks learning residual functions with respect to their inputs.
Huang et al. \cite{huang2016deep} proposed to train ResNet with a random subset of layers dropped and bypassed with ResNet's skip connections, yielding a network with stochastic depth.
This technique acts as a regularizer, although it might work the best for very deep networks.
For shallower networks (like ones considered in this paper)  it might be a too strong form of regularization.

In this paper, we propose a method for automatic ICAC segmentation.
Our method is based on a deeply supervised 3D FCN.
To ease the optimization, we introduce a simple problem-specific modification of the network's objective that emphasizes important image regions.
To further increase the generalization capacity of the network, we propose to combine dropout and ResNet by inserting dropout layers into the residual blocks.
We investigate the importance of ResNet's skip connections and the position of the dropout layers in blocks.

\section{Methods}

Our architecture is described in detail in Section \ref{sec:arch}. In the following subsection we explain deep supervision and how we adapt the objective to our problem. In the last subsection we explain dropout ResNets.

\subsection{Architecture}
\label{sec:arch}

\begin{figure}[!t]
\centering
\includegraphics[width=0.9\textwidth]{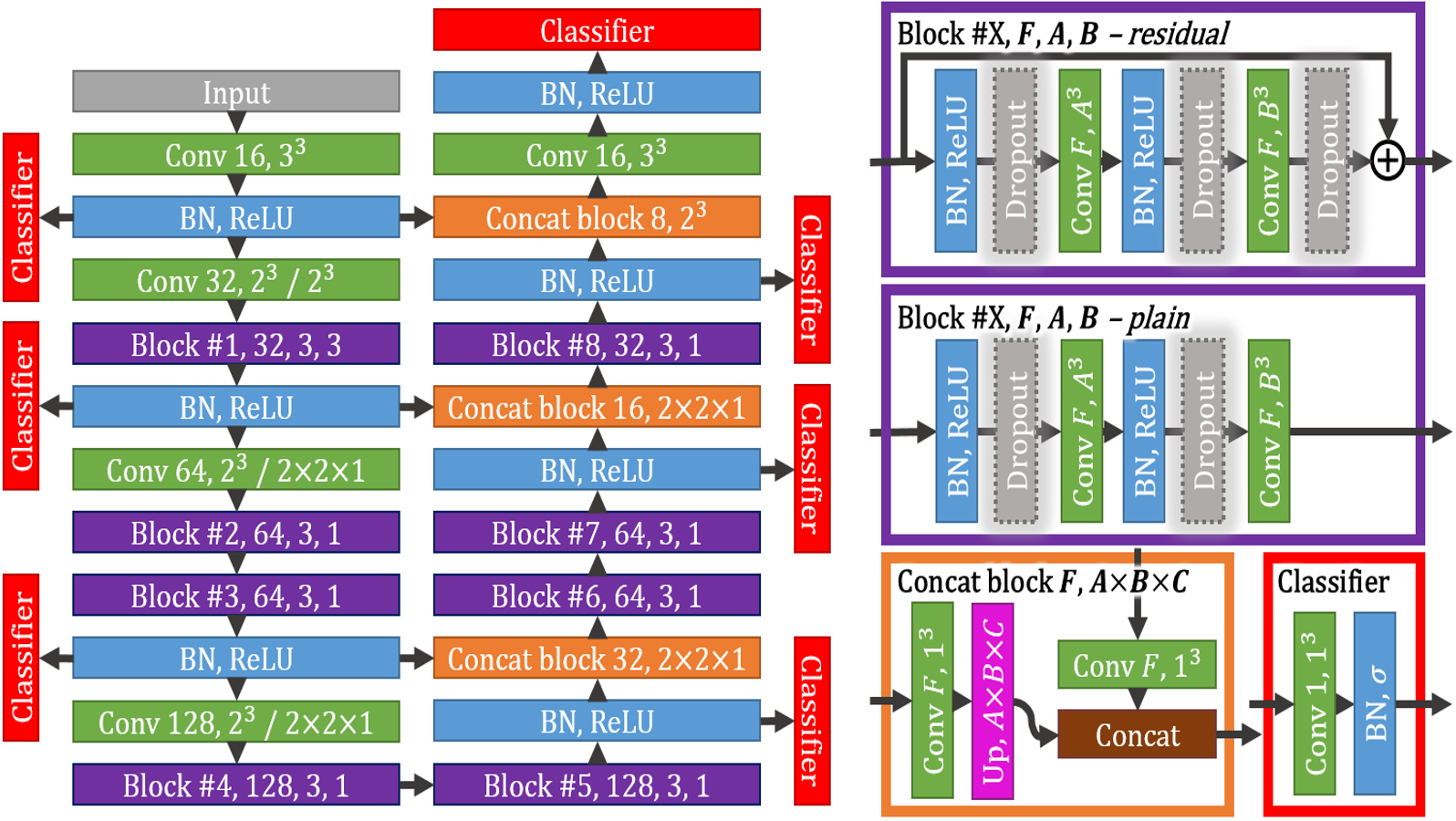} %
\caption{The architecture of our network. Green ``Conv'': (strided) convolutional layers with parameters indicated as ``\{output feature number\}, \{kernel size\} / \{stride\}''. Pink ``Up'': upsampling layers. Blue ``BN, ReLU'' or ``BN, $\sigma$'': batch normalization (BN) and ReLU/sigmoid activation. Brown ``Concat'': concatenation along feature dimension. Grey ``Dropout'': possible positions of dropout layers. For the sake of simplicity, cropping layers needed to match the dimensions of inputs to concatenation and summation layers are omitted.} %
\label{fig:arch} %
\end{figure} %

Our base architecture is depicted in Fig. \ref{fig:arch}.
We use valid convolutions to avoid undesired border effects.
Strided convolution is used for downsampling.
All convolutions, down- and upsampling operations are 3D. We downsample less along the longitudinal axis, because a very large receptive field along that dimension is not necessary.
Batch normalization (BN) \cite{ioffe2015batch} layers are added before every activation layer, to improve the convergence speed and regularize the network.
The receptive field of our network is $85 \times 85 \times 37$ voxels.
The number of features and layers is chosen to fit the available GPU memory (8 GB).

In this paper, we experiment with residual and non-residual, or ``plain'', architectures.
We obtain a ResNet or a plain variant of our architecture by choosing a corresponding block (violet blocks in Fig. \ref{fig:arch}).
A residual block takes an input $x$ and outputs $\mathcal{H}(x)$, with its layers computing a residual function $\mathcal{F}(x) = \mathcal{H}(x) - x$.
The layer order in our residual blocks is the same as the one proposed in \cite{he2016identity}.

\subsection{Deep Supervision and Objective Function}

Our network has six auxiliary classifiers placed on top of several intermediate layers (Fig. \ref{fig:arch}).
The training objective is the sum of the loss associated with the final classifier and a weighted sum of the auxiliary classifier losses: $\mathcal{L}^{total}(W, \textbf{w}) = \mathcal{L}(W; \mathcal{X}, \mathcal{Y}) + \sum_i{a_i\mathcal{L}_i(W, w_i; \mathcal{X}, \mathcal{Y})}$, where $\mathcal{X}$ is a collection of input voxels, $\mathcal{Y}$ is corresponding ground truth labels, and $W$ and $\textbf{w} = [w_1, ..., w_6]$ are the weights of the main network and auxiliary classifiers, respectively.

Every $\mathcal{L}_i$ is the sum of cross-entropy losses measuring the mismatch between the network output and the ground truth for voxels above the clinical calcification threshold of 130 HU.
We exclude below-threshold voxels from supervision in order to simplify the optimization problem and restrict the focus of the network to distinguishing between ICAC and the most difficult negatives -- bones.

\subsection{Dropout in Residual Networks}

Dropout is a technique in which activations of a randomly selected subset of neurons are set to zero during training.
During testing, no neurons are dropped, but the weights of the network are decreased to account for the resulting increase in total activation.
Dropout reduces co-adaptation between neurons, and hence yields a regularization effect \cite{srivastava2014dropout}.

However, dropout layers cause a complete loss of a subset of their input features, and thus reduce representation capacity of plain models during training, which may harm their performance.
For this reason, we believe that a combination of dropout with a ResNet architecture is an interesting alternative.
We obtain a dropout variant of our architecture by placing dropout layers inside its blocks (at most one per block).
Fig. \ref{fig:arch} shows all possible positions of dropout layers in plain and residual blocks (dropout cannot precede BN).
Unlike its plain counterpart, the residual dropout version of our network always maintains a full set of features, due to skip connections circumventing dropout layers.

The difference between stochastic depth \cite{huang2016deep} and our approach is the aggressiveness of dropout.
In stochastic depth, either none of the features of a block's last layer are randomly set to zero, or all of them, which effectively shortens the network depth during training.
In our method, either block's convolutional layers get a noisy input (if dropout is placed before convolution), or a block's last layer is corrupted by noise (with dropout before addition).
A middle ground between stochastic depth and our dropout ResNet could be a dropout ResNet in which the standard dropout layers (as in \cite{srivastava2014dropout}) are replaced by more aggressive SpatialDropout \cite{tompson2015efficient} layers, which  randomly drop entire feature maps.


\section{Dataset, Preprocessing and Network Training}

Our dataset consists of 1882 non-contrast-enhanced CT images reaching from the aortic arch to the intracranial vasculature.
The images were annotated by two trained observers who indicated regions of interest (ROI) with visible calcification on all image slices.
ICAC lesions are easily obtained from these annotations by thresholding at 130 HU.
The in-plane resolution of scans is $\SI{0.23}{mm} \times \SI{0.23}{mm}$ and the slice thickness is $\SI{1}{mm}$.
We registered the images rigidly to a single reference image and cropped them along the longitudinal axis so that they contain only the intracranial part of the carotid artery.
Finally, we downsampled axial slices (roughly twice) so that their spatial resolution matches that of the longitudinal axis.
The final image size is $240 \times 240 \times 100$ voxels.

We assigned every image randomly to the training, validation and test sets of sizes 632, 250 and 1000 respectively.
Due to the GPU memory limitations, we trained our networks on mini-batches of one patch of $178 \times 178 \times 98$ voxels.
During testing, a network was applied on the images patch-wise.
The output patches were tiled and averaged (at locations of overlap) to yield a segmentation of a whole scan.
The final segmentation was obtained by removing the voxels below 130 HU.
The only kind of data augmentation used was flipping along the frontal axis.
The network was trained with stochastic gradient descent with an initial learning rate of 0.1, which was reduced 10 times after epoch 10.
The momentum was increased from 0.9 to 0.99 after the same epoch.
The initial weight of positive voxels (ICAC) in the objective was set to 1000 (the approximate ratio between positives and negatives) and reduced to 1 after epoch 5.
The learning rate, momentum and voxel weighing schedules were chosen to yield fast and smooth training loss decay.
The weights $a_i$ of auxiliary classifier losses were initialized at 1 and decayed linearly to 0 over the course of 50 epochs.
Dropout layers were inserted in blocks 4-8 with dropout probabilities of [0.3, 0.3, 0.4, 0.4, 0.5].

All hyperparameters were selected based on the experiments on the training and validation sets, prior to the evaluation on the test set.
The final network weights were chosen based on the validation loss.

\section{Results and Discussion}
\label{sec:results}

\begin{table}[!t]
\setlength{\tabcolsep}{3.7pt}
\centering
\caption{
The contribution of different techniques to the performance.
Absolute Dice measures the overlap between the network segmentations and the ground truth without averaging over the images.
Quarters are defined by ordering the images by increasing ICAC volume, and partitioning into four equally-sized groups.
The last column reports the significance of the improvement over the previous row computed with a paired t-test.
}
\begin{tabular}{c c c c c c c c}
\hline\rule{0pt}{12pt}
\multirow{2}{1.8cm}{ \centering Experiment}
&
\multirow{2}{1.4cm}{ \centering Absolute Dice}
&
\multirow{2}{1.8cm}{ \centering Dice Mean and SD}
&
\multicolumn{4}{p{3.3cm}}{ \centering Mean Dice per Quarter}
&
\multirow{2}{1.1cm}{ \centering T-test $P$}
\\
 & & & 25 & 50 & 75 & 100 & \\
\hline\rule{0pt}{12pt}
plain 3D FCN				&	80.1	&	$69.1 \pm 21.8$ & 52.6 & 67.9 & 75.3 & 80.9 & - \\
+ deep supervision			&	83.1	&	$72.9 \pm 21.9$	&	58.6 & 71.3 & 77.7 & 84.2 & $<10^{-22}$ \\
+ $>130$ HU objective			&	84.8	&	$75.1 \pm 22.3$	&	60.1 & 73.9 & 80.4 & 86.3 & $<10^{-9}$ \\
+ best dropout ResNet			&	85.0		&	$76.2 \pm 20.9$	&	62.8 & 75.3 & 80.4 & 86.5 & 0.0044 \\
\hline
\end{tabular}%
\label{tab:general}%
\end{table}

We evaluate the effect of our techniques on the performance by progressively adding them to our baseline network: a plain 3D FCN, without deep supervision and dropout, and with supervision of all voxels.
Table \ref{tab:general} presents the results of the evaluation on the test set.
Deep supervision increased the Dice substantially.
However, unlike \cite{lee2015deeply} we did not observe an increased convergence speed.
We suspect that this might be linked to BN already speeding up the optimization.

Removing below-threshold voxels from supervision improved the Dice overlap and convergence speed: the network reached 75\% mean training Dice at epoch 24, whereas the network with supervision of all voxels did so at epoch 57.
One explanation could be that removing a large part of the voxels from the objective made the optimization problem easier (i.e., there were fewer constraints to satisfy).
Another explanation for the improvement in Dice could be that our objective emphasizes difficult negatives, whereas in the objective that supervises all voxels those negatives constitute only around a fifth of all negatives.

The best performance was achieved by turning the network into a ResNet and adding dropout layers into the residual blocks before addition.
Interestingly, unlike the other methods, dropout increased the Dice of images with smaller ICAC volumes much more than the Dice of the other images (see the Dice per quarter in Table \ref{tab:general}).
This is because dropout improved the performance more for smaller and lower intensity (close to 130 HU) lesions, which occur more often on images with a lower ICAC volume.
This might happen because networks with dropout use more information from neighboring areas, which can be helpful for smaller and lower intensity lesions, because downsampling might reduce their intensity to a value below the calcification threshold.

\begin{table}[!t]
\setlength{\tabcolsep}{3pt}
\centering
\caption{
Results of the experiments with plain and residual architectures and different dropout positions.
Every network is a deeply supervised 3D FCN with supervision of only $>130$ HU voxels.
Positions correspond to those in Fig. \ref{fig:arch}.
The last column indicates the significance of the improvement over the plain non-dropout network.
}
\begin{tabular}{c c c c}
\hline\rule{0pt}{12pt}
Experiment & Absolute Dice & Dice Mean and SD & T-test $P$\\
\hline\rule{0pt}{12pt}
Plain non-dropout					&	84.8	&	$75.1 \pm 22.3$ & - \\
Plain dropout						&	83.0	&	$70.7 \pm 24.7$ & $< 10^{-17}$\\
ResNet + dropout before first conv.	&	84.9	&	$75.7 \pm 21.4$ & 0.0694 \\
ResNet + dropout before second conv.	&	84.3	&	$76.0 \pm 19.8$ & 0.0261 \\
ResNet + dropout before addition		&	85.0	&	$76.2 \pm 20.9$ & 0.0044 \\
\hline
\end{tabular}%
\label{tab:dropout}%
\end{table}

We evaluated the importance of skip connections and the positioning of dropout layers in the ResNet blocks.
The results are presented in Table \ref{tab:dropout}.
Dropout layers in the plain architecture decreased the Dice.
We believe this is because our network has a rather small number of features for our problem complexity and dropout reduced the representation capacity of the model too much.
This is supported by the substantially reduced performance we observed for networks with a smaller number of features (results not shown).
Stochastic depth \cite{huang2016deep} and ResNet with SpatialDropout also reduced the performance (results not shown).
We suspect it is because of the same reason: these techniques are too strong regularizers for a network of our size applied to our problem (e.g., 100 layers of \cite{huang2016deep} vs. ours 24).
In contrast, our dropout ResNet does not cause a significant reduction in the representation capacity or expressivity of the model, and still exerts a regularizing effect as it reduces co-adaptation between neurons.

Placing dropout before addition produced a slightly higher Dice than placing it before one of the two convolutions ($P$-values $0.106$ and $0.328$).
When dropout is placed before convolutional layers, these layers can compensate (to some extent) the information loss induced by dropout before it is passed to a next block, whereas in the other case, the corrupted output is passed to the next block and even further with skip connections.

\begin{figure}[!t]
\centering
\includegraphics[width=1.0\textwidth]{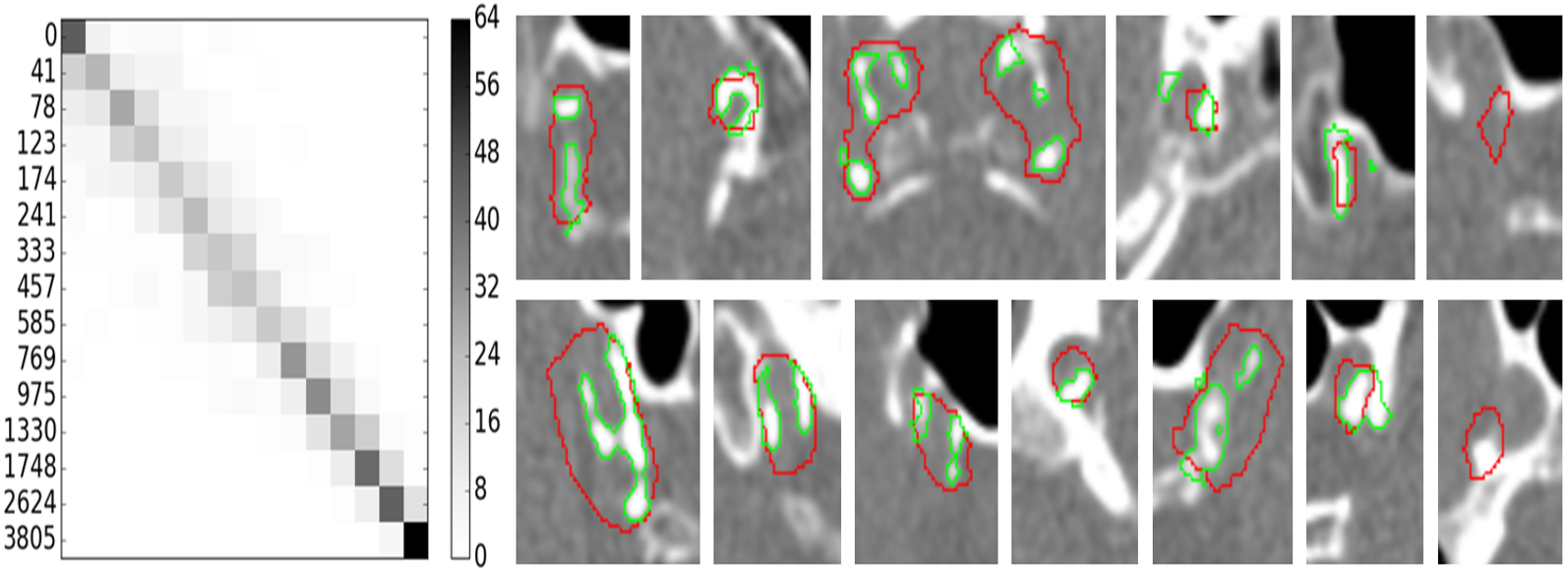}%
\caption{\emph{Left}: a 2D histogram of automatically computed ($y$ axis) and ground truth ICAC volumes ($x$ axis). Every bin has the same number of images. \emph{Right}: example detections. Red: ground truth ROIs. Green: network segmentations.}%
\label{fig:det}%
\end{figure}

Fig. \ref{fig:det} shows examples of detections performed by our best network.
Our network can accurately segment ICAC even when it is adjacent to bone, although sometimes it still captures a part of the bone.
The toughest examples for detection, responsible for over 80\% of missed lesions, were very small lesions and lesions with an intensity close to 130 HU.
Lesions adjacent to bones were also over-represented among the false negatives, but to a substantially lesser extent.

The intraclass correlation coefficient (ICC) between the automatically estimated ICAC volumes and the ground truth for our best network is 97.7\%. (See Fig. \ref{fig:det}.)
This is quite close to the interrater agreement with ICC = 99\% that we computed on 50 images from our dataset.

\section{Conclusion}

We presented a method for automatic segmentation of ICAC in non-contrast-enhanced CT.
We introduced several modifications to a plain 3D fully convolutional network, namely: supervision of hidden layers, dropout combined with residual architecture, and a problem-specific adaptation of the objective function restricting the focus on the most relevant structures.
Every modification resulted in a statistically significant improvement, totaling to 7.1\% increase in the mean Dice.
The agreement between our best network ICAC volume estimations and the expert estimations is close to the interobserver agreement for our dataset.
We believe our method has a potential for application on large-scale epidemiological studies on ICAC.

%
%

\bibliography{bibl}


\end{document}